\documentclass[preprint,12pt]{elsarticle}

\usepackage[utf8]{inputenc}
\usepackage{xcolor}
\usepackage{amssymb}
\usepackage{float}
\usepackage{amsmath}
\usepackage{hyperref}
\usepackage{graphicx}
\usepackage{caption}
\usepackage{subcaption}
\usepackage{booktabs} 
\usepackage{algorithm,algpseudocode}
\usepackage{array}
\usepackage{hhline}
\usepackage{appendix}
\usepackage{listings}
\usepackage{natbib}
\usepackage[displaymath, mathlines]{lineno}
\setcitestyle{authoryear,open={(},close={)}} 

\begin{document}

\begin{frontmatter}

\title{Deep Learning for Sea Surface Temperature Reconstruction under Cloud Occlusion}

\author[1]{Andrea Asperti}
\author[2]{Ali Aydogdu}
\author[1]{Angelo Greco}
\author[1]{Fabio Merizzi}
\author[2]{Pietro Miraglio}
\author[2]{Beniamino Tartufoli}
\author[1]{Alessandro Testa}
\author[2,3]{Nadia Pinardi}
\author[2,3]{Paolo Oddo}

\affiliation[1]{organization={University of Bologna, Department of Informatics - Science and Engineering (DISI)},
            addressline={Via Mura Anteo Zamboni 7}, 
            city={Bologna},
            postcode={40126}, 
            state={},
            country={Italy}}        

\affiliation[2]{{CMCC Foundation - Euro-Mediterranean Center on Climate Change},
            addressline={Viale Carlo Berti Pichat 6/2}, 
            city={Bologna},
            postcode={40127}, 
            state={},
            country={Italy}}      
            
\affiliation[3]{organization={University of Bologna, Department of Physics and Astronomy (DIFA)},
            addressline={Viale Carlo Berti Pichat 6/2}, 
            city={Bologna},
            postcode={40127}, 
            state={},
            country={Italy}}

\begin{abstract}
Sea Surface Temperature (SST) reconstructions from satellite images affected by cloud gaps have been extensively documented in the past three decades. Here we describe several Machine Learning models to fill the cloud-occluded areas starting from MODIS Aqua nighttime L3 images. To tackle this challenge, we employed a type of Convolutional Neural Network model (U-net) to reconstruct cloud-covered portions of satellite imagery while preserving the integrity of observed values in cloud-free areas. We demonstrate the outstanding precision of U-net with respect to available products done using OI interpolation algorithms. Our best-performing architecture show 50\% lower root mean square errors over established gap-filling methods.
\end{abstract}


\end{frontmatter}

\section{Introduction}

Sea Surface Temperature (SST) is one of the essential climate variables for understanding and modeling the Earth's climate system. Air-sea heat fluxes, depending on the sea surface and air temperature difference, are a major driver of the atmosphere-ocean coupled dynamics. Satellite SST products, interpolated on regular grids and cloud gaps filled \citep{reynolds2002improved, konik2019operational}, continue to be used today to force atmospheric model analyses and reanalyses \citep{donlon2012operational}. The influence of high-resolution sea surface temperatures (SST) on the accuracy of atmospheric reanalysis has been demonstrated \citep{parfitt2017impact}, along with the role of SST fronts in driving climate variability \citep{larson2024signature}. Therefore, we aim to investigate innovative methods to fill cloud-occluded areas without introducing smoothing effects on the cloud free pixels.

The SST satellite measurement is typically made by sensing the ocean radiation in many wavelengths within the near infrared and microwave parts of the electromagnetic spectrum. One difficulty in getting a global product using only infrared imagery is that there are large regions with clouds \citep{wylie2005trends}, and reconstructing SST under cloud occlusion conditions is a challenging and lively research topic. Traditional statistical techniques, such as Optimal Interpolation \citep[OI;][]{bretherton1976technique}, Empirical Orthogonal Function \citep[EOF;][]{alvera2011data}, and other similar techniques \citep{jung2022high, catipovic2023reconstruction}, fill missing values based on spatial/temporal correlations between observed SST points. The present-day SST Level 4 (L4) product from Copernicus Marine Service is produced by a spacetime OI scheme \citep{nardelli2013high}. 

A problem of statistical objective mapping techniques is that they often struggle to resolve fine-scale features, resulting in smooth reconstructions \citep{chin2017multi, fablet2017data, barth2020dincae}. These techniques typically assume linearity, which limits their ability to capture the complex dynamics of SST fields. One of the most advanced SST multi-sensor reconstructions \citet{chin2017multi} currently employs adaptable time windows to fill image gaps caused by cloud cover. However, this approach has several issues, including the potential for mesoscale signals beneath the clouds to be biased by older data from several days earlier, leading to contamination of the reconstructed area.

Consequently, there has been growing interest in recent years in applying deep learning techniques to address this challenge. Convolutional autoencoders, such as DINCAE, have been successfully used to reconstruct SST and chlorophyll concentrations \citep{barth2020dincae, barth2022dincae, han2020application}, and vision transformers trained with a masked autoencoder approach  have also been explored \citep[MAESSTRO;][]{goh2024maesstro}. A recent survey of statistical and AI-based reconstruction methods for oceanographic data is available in \citet{catipovic2023reconstruction}.

Image completion -or image inpainting- is a well-researched area in image processing \citep{iizuka2017globally, liu2018image, peng2021generating, wan2021high, zheng2022bridging, jain2023keys}, with various techniques successfully applied across different domains. In this article, we evaluate some of these image completion techniques for reconstructing SST under cloudy grid points. The method is applied to SST Level 3 (L3) images at the resolution of 4 km from the MODIS Aqua satellite and infrared sensor; particular attention is given to the model configuration and the calibration of the parameters. After this, we apply the best algorithm to another L3 operational product from Copernicus Marine Service.

\begin{figure}[th]
    \centering
    \includegraphics[width=0.9\linewidth]{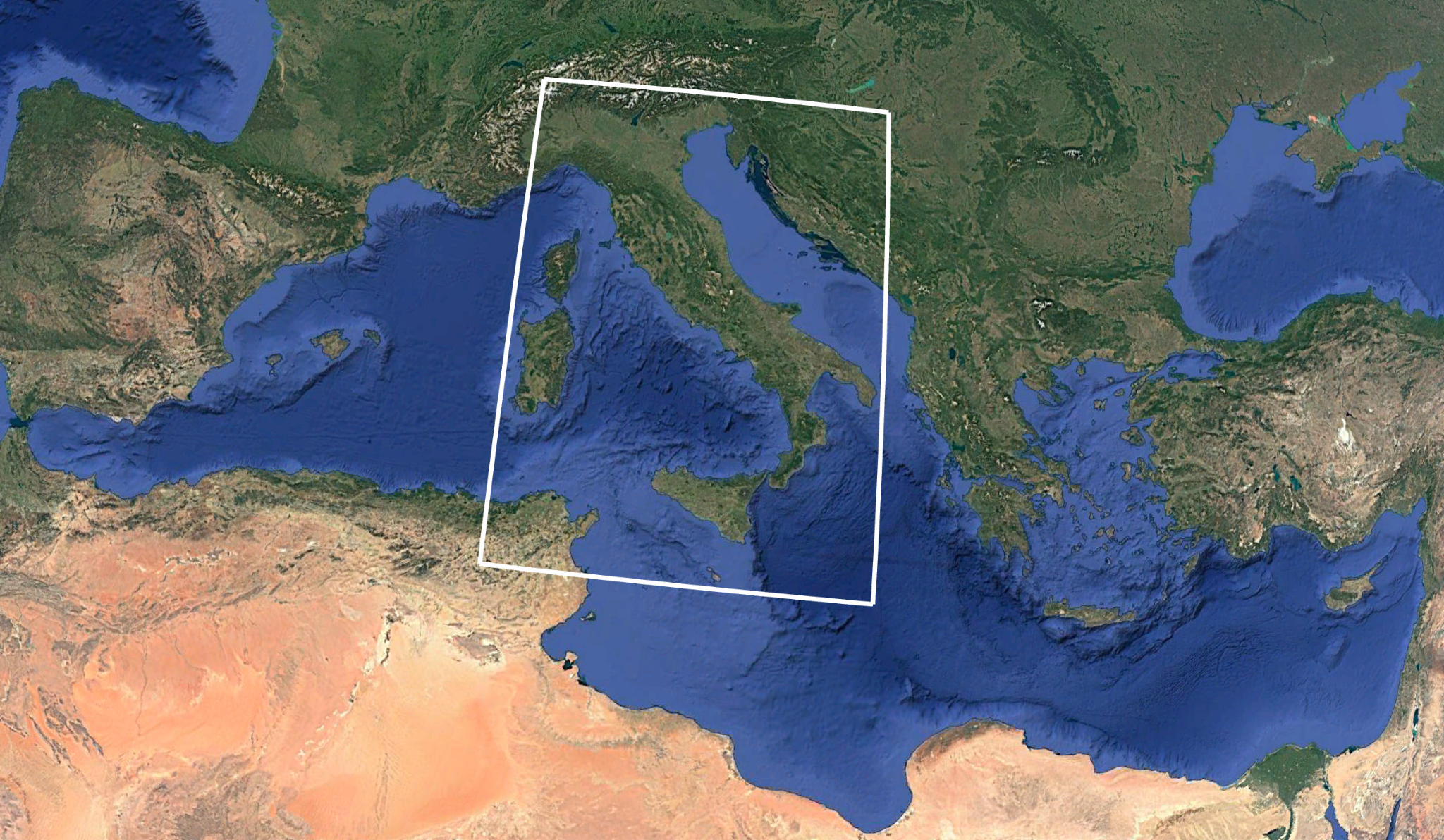}
    \caption{The white polygon describes the region of our investigation, with latitude between 35.33$^\circ$ and 46.0$^\circ$E and longitude between 7.92$^\circ$ and 18.58$^\circ$N.}
    \label{fig:area}
\end{figure}

The analysis is done on a region comprising the Italian Seas, described in the white box of Figure~\ref{fig:area}. The area covers 256x256 observation points with a latitude between 35.33$^\circ$ and 46.0$^\circ$E and a longitude between 7.92$^\circ$ and 18.58$^\circ$N, for a total domain extension of 1020 x 1020 square kilometers. The reconstruction obtained by our model was validated against the current L4 reconstructions from Copernicus Marine Service \citep{nardelli2013high, pisano2022new} and the recent DINCAE model \citep{barth2020dincae, barth2022dincae}, testifying in both cases a notable performance improvement.

The article is structured in the following way. In Section~\ref{sec:dataset}, we describe the satellite dataset used for the initial training of the neural network. This section also contains an investigation of data, comprising gradients (subsection~\ref{sec:gradients}), persistence (subsection~\ref{sec:persistence}), and climatology (subsection~\ref{sec:climatology}). The methodology is explained in Section~\ref{sec:models}, where we introduce the main classes of models investigated in this article, namely U-Net models (subsection~\ref{sec:unet}) and Visual Transformers (subsection~\ref{sec:vit}). In this section we also discuss the artificial cloud generator (subsection~\ref{sec:generator}), and training (subsection~\ref{sec:training}). The model intercomparison and selection is illustrated in Section~\ref{sec:results}. Section~\ref{sec:copernicus} applies the trained network to an operational L3S dataset for comparison with established products. We discuss the results in section~\ref{sec:discussion} and draw the conclusions in section~\ref{sec:conclusions}.

\section{Satellite Data Set for algorithm development}
\label{sec:dataset}

The training SST time series is taken from the MODIS dataset (available at \url{https://podaac.jpl.nasa.gov/dataset/MODIS_AQUA_L3_SST_THERMAL_DAILY_4K_NIGHTTIME_V2014.0}, last accessed Sept 2024). Data are acquired by the Moderate-resolution Imaging Spectroradiometer \citep[MODIS;][]{werdell2013generalized} on board the NASA TERRA and AQUA satellite platforms, launched in 1999 and 2002 respectively. For our investigation, we used the daily products at 4 km spatial resolution relative to nighttime passes. We reconstructed the MODIS-AQUA data, using MODIS-TERRA for validation purposes. The MODIS-AQUA data set contains SST values with missing data due to cloud occlusions. All nighttime measurements from 7/4/2002 to 12/31/2023 were used. Part of the data, from 7/4/2021 to 12/31/2023, was used for testing purposes. The data set also contains quality flags for each grid point. The flags go from 0 (best) to 5 (worst). We used only values of quality 0, 1 and 2, amounting respectively to 79\%, 21\% and 0.2\% of grid points for our region of interest. 

\subsection{Data set analysis}

\begin{figure}[h]
    \centering
    \includegraphics[width=0.5\linewidth]{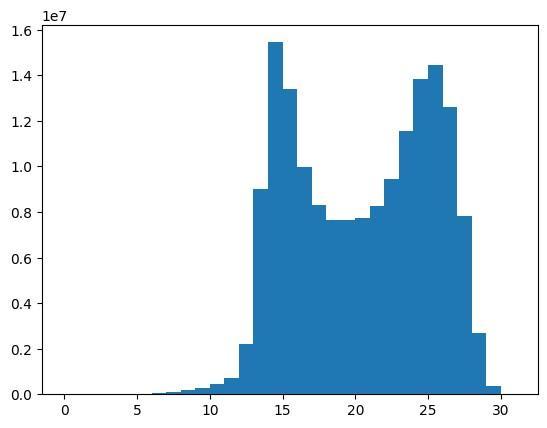}
    \caption{Histogram relative to the distribution of MODIS-AQUA nightly temperatures, cumulative overall spatial positions and all years.}
    \label{fig:SST-histogram}
\end{figure}

In this section, we investigate the datasets, pointing out a few critical aspects of collected data, typically due to problems of the signal in proximity of the coast, or at the border of clouds. The minimum, maximum and average values for all nighttime SST grid values from MODIS-AQUA are 0.1, 31.1 and 20.5$^\circ$C respectively. 
The Mediterranean Sea SST never falls below 5$^\circ$C, thus the minimum temperature is likely due to cloud borders incorrectly associated to seawater by the cloud detection algorithm. These outliers are very few in number and have negligible impact on training or evaluation.
As expected for the Mediterranean Sea, a large amplitude seasonal cycle is present (Fig.~\ref{fig:SST-histogram}) and a suitable seasonal climatology should be computed, as described in the section below.

\subsection{SST Gradients}
\label{sec:gradients}

Here, we define gradients as differences in SST grid points, both in time and space. The gradients are defined as the difference between two consecutive nighttime values in the same grid point and the difference between two consecutive grid points every night (4 km grid spacing). Table~\ref{tab:fluctuations} presents statistics related to these gradients.

\begin{table}[h]
\centering
\begin{tabular}{c|c|c|c|c|c|c|c|c|}
\hhline{~--------}
&\multicolumn{4}{|c|}{temporal axis} & \multicolumn{4}{c|}{spatial axis}\\\hhline{~--------}
&max & avg max & mean & std & max & avg max & mean & std \\\hline
\multicolumn{1}{|c|}{night} & 8.2 & 4. & 0.4 & 0.4 & 8.1 & 3.5 & 0.2 & 0.2 \\\hline
\end{tabular}
\caption{Statistics on the spatial and temporal sea surface temperature gradients, measured as differences between consecutive nights and neighboring points in space within the same night calculated on all data from 2002 to 2023.}
\label{tab:fluctuations}
\end{table}

\begin{figure}[h]
    \centering
    \includegraphics[width=\textwidth]{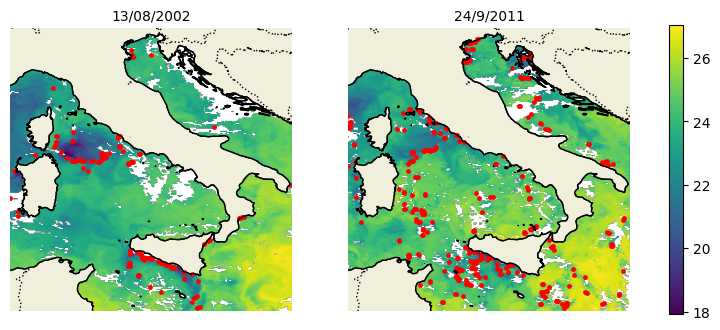}
    \caption{Large Spatial Gradients ($>1.5^\circ$C) relative to different days of the year. Units are $^\circ$C}
    \label{fig:gradients}
\end{figure}

Spatial gradients are generally low, with an average fluctuation around 0.1/0.2~$^\circ$C. However, these fluctuations are not uniformly distributed: larger gradients are typically observed near coastal areas (see Fig.~\ref{fig:gradients}), particularly along the western coast of the Adriatic Sea and the southern Sicilian coast.

In some days, large gradients are found in the open ocean around large scale oceanic features. An example is given in Fig.~\ref{fig:gradients}, where the maximum gradients are around the southern border of the Northern Tyrrhenian cyclonic gyre \citep{pinardi2015mediterranean}, east of the Strait of Bonifacio. In other cases, the extreme gradient values are around the cloud borders, due to satellite cloud removal accuracy. The mean temporal variation from one day to the next is around 0.4~$\circ$C (Table~\ref{tab:fluctuations}).

Table~\ref{tab:frequency_of_grads} shows the percentage of points with extreme gradients, considering thresholds between 1 and 3$^\circ$C. There are almost no nighttime gradients larger than 1$^\circ$C. 

\begin{table}[h]
\centering
\begin{tabular}{r|c|c|c|c|c|}
\hhline{~-----}
& $> 1^\circ$ & $> 1.5^\circ$ & $> 2^\circ$ & $ > 2.5^\circ$ & $ > 3^\circ$ \\\hline
\multicolumn{1}{|c|}{night:time} & 1.1\% & 1\% & 0.9 \% & 0.8\% & 0.7\% \\\hline
\multicolumn{1}{|c|}{night:spatial} & 0.3\% & 0.1\% & - & -  & - \\\hline
\end{tabular}
\caption{Frequency of high spatial and temporal gradients in MODIS-AQUA data in the region of interest and for the time period 2002-2023. In the spatial case, the “gradient” refers to the difference in SST between two neighboring grid points, 4 km apart; similarly, for the temporal case, it is the difference in SST between two consecutive days at a given spatial position}
    \label{tab:frequency_of_grads}
\end{table}

\subsection{Filling cloudy pixels with temporal interpolation}
\label{sec:persistence}

As in many classical reconstructions of SST below the clouds \citep{chin2017multi}, it is common to use for each target night a temporal sequence of a few consecutive days to temporally extrapolate/interpolate SST values on the cloudy pixel. This approach can be seen as a form of persistence filling algorithm, where we use the closest available data to fill gaps.
Unfortunately, the approximation provided by data from previous days is very inaccurate. Table~\ref{tab:persistence} shows the difference for nighttime data, expressed in terms of both Mean Absolute Difference (MAD) and Root Mean Square Difference (RMSD), for increasing temporal gaps, ranging from 1 to 5 days. The latter is the temporal window used with the current SST reconstruction methods \citep{catipovic2023reconstruction, chin2017multi}. The difference is measured as an average over sea locations that are uncontaminated by clouds on both days. The RMSD and MAD increase very rapidly, and the quality of the reconstruction using data from previous days is doubtful.

\begin{table}[h]
    \[
    \begin{array}{l|c|c|c|c|c|}
    \hhline{~-----}
            & 1d & 2d & 3d & 4d & 5d\\\hline
        \multicolumn{1}{|c|}{\textbf{MAD}} & 0.4^\circ & 0.5^\circ & 0.6^\circ & 0.7^\circ & 0.8^\circ \\\hline
        \multicolumn{1}{|c|}{\textbf{RMSD}} & 0.5^\circ & 0.7^\circ & 0.8^\circ & 1^\circ & 1.1^\circ\\\hline
    \end{array}
    \]
    \caption{Average distance, in terms of Mean Absolute Difference (MAD) and Rooted Mean Squared Difference (RMSD).}
    \label{tab:persistence}
\end{table}

According to our experiments, reported in Section~\ref{sec:results}, considering temporal sequences longer than  4 consecutive days bring no improvement to the reconstruction.

\subsection{Seasonal Climatology}
\label{sec:climatology}

As it is clear from Fig.~\ref{fig:SST-histogram}, climatology has a large seasonal cycle. In statistical analysis, it is important to subtract the quasi-periodic signals in the time series, such as the seasonal cycle. Thus, we compute the seasonal climatology as the time mean SST across all the 21 years dataset. Anomalies are then calculated by subtracting from each day the seasonal climatology.

The computed daily climatology still suffers from small time gaps where SST values are absent for a particular day of the year, due to the presence of clouds. To fill the climatology gaps, we use interpolation and specifically a Gaussian blur, which uses nearby spatial values to estimate missing ones. The Gaussian function gives more weight to closer values, while gradually reducing the influence of distant points. The algorithm is briefly described in Appendix A. We also considered an “unbiased” version of the climatology, where we adjust the mean SST of the baseline towards the observed daily temperature from non-cloudy pixels.

\begin{figure}[h] 
\centering
   \includegraphics[width=\linewidth]{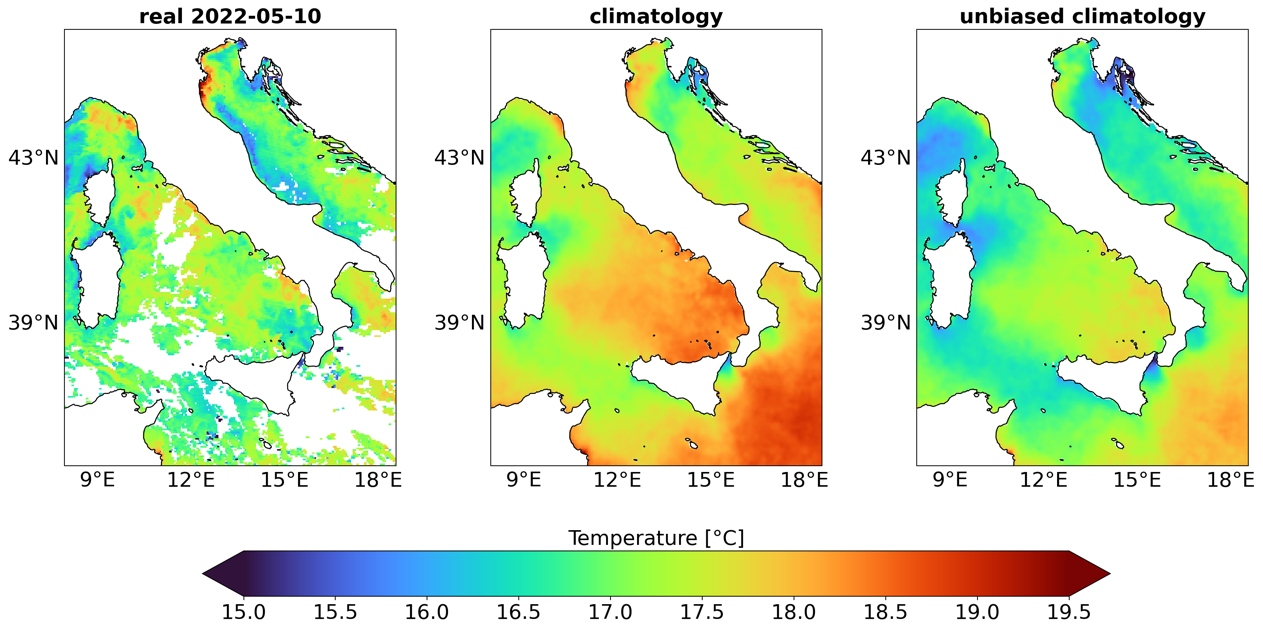}
   \caption[Climatology]{On the left, the nighttime SST data from a sample day, in this case 10/05/2022; in the middle, the climatology for May 10; on the right, the climatology adjusted (shifted) to the mean of the specific day.}
   \label{fig:climatology} 
\end{figure}

Figure~\ref{fig:climatology} shows an example of daily climatology and its unbiased version. As is evident from the figure, both the daily climatology and its unbiased version have limitations because the climatology can differ significantly from the specific day under consideration. The unbiased version is usually closer to reality but still not capable of capturing the single day values. For our experiments, we trained the model to learn the residual information with respect to climatology.

\section{Models}
\label{sec:models}
We tested several different neural network architectures, including variants of U-Net \citep{ronneberger2015u}, Visual Transformers \citep[ViT;][]{dosovitskiy2020image}, and Diffusion Models \citep{song2020denoising, ho2020denoising}. Each model was evaluated across a wide range of configurations, varying input dimensions, the number of channels, network depth, and incorporating specific modules such as attention layers or inception modules. So far, we have not achieved satisfactory results with diffusion models, so we will not report on those results.

Special attention was given to determining the optimal size of the geographical area under investigation. Experimentally, we found that splitting the original 256x256 region into four smaller areas, each 128x128 in size, and training four separate models resulted in better performance. Another key focus of the experimentation was determining the appropriate length of the temporal sequence of consecutive days to be used as input to the model.

In this section, we briefly introduce the two main classes of models: UNet and ViT.

\subsection{U-Net}
\label{sec:unet}

The U-Net is a type of convolutional neural network (CNN) originally designed for biomedical image segmentation. Its architecture is structured as a U-shaped network, consisting of two main parts: the contracting path (encoder) and the expansive path (decoder). The encoder progressively reduces the spatial dimensions of the input image through convolutional and pooling layers, capturing increasingly abstract and high-level features. The decoder, in contrast, upsamples the feature maps to the original input size, allowing for precise localization in the reconstructed output. 
The number of downsampling layers and their respective number of channels are key hyperparameters of the network. For example, a U-Net with the structure [64, 128, 256, 512] refers to a model with three downsampling layers that progressively halve the spatial dimensions while increasing the depth from the initial 64 channels to 128, 256, and 512 channels, respectively.

A key innovation of U-Net is the use of skip connections between corresponding layers in the encoder and decoder. These connections transfer high-resolution feature maps from the encoder to the decoder, allowing the model to combine both coarse and fine-grained information during reconstruction. This design makes U-Net highly effective for tasks where detailed output is essential. The U-Net was also already used by \cite{barth2020dincae,barth2022dincae} in its cloud filling algorithm using two U-Net in sequence.

\begin{figure}[ht]
\centering
\includegraphics[width=\textwidth]{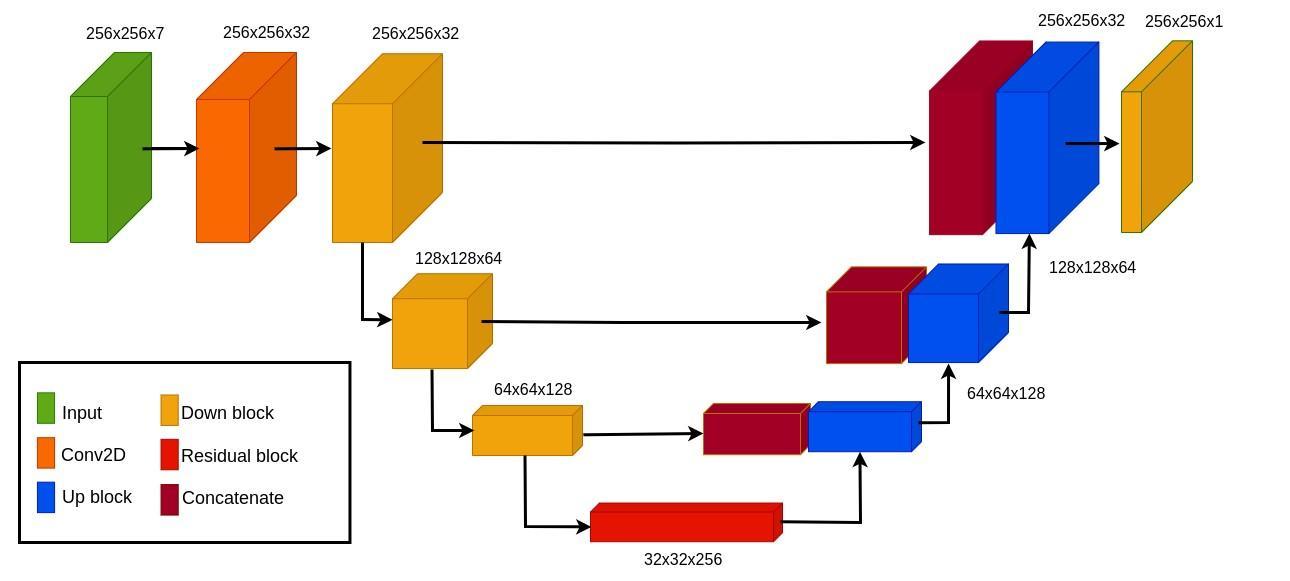}
\caption{Basic U-Net. In our terminology, this U-Net has a [32,64,128,256] structure, meaning that it is composed of three downsampling blocks progressively halving the spatial dimension, and increasing the channel dimension to 64, 128 and 256. The initial spatial dimension is 256x256. The initial number of channels is 7, corresponding to three input days with the associated masks and the land-sea mask.
}\label{fig:unet}
\end{figure}

\begin{figure}[H]
\centering
\includegraphics[width=\textwidth]{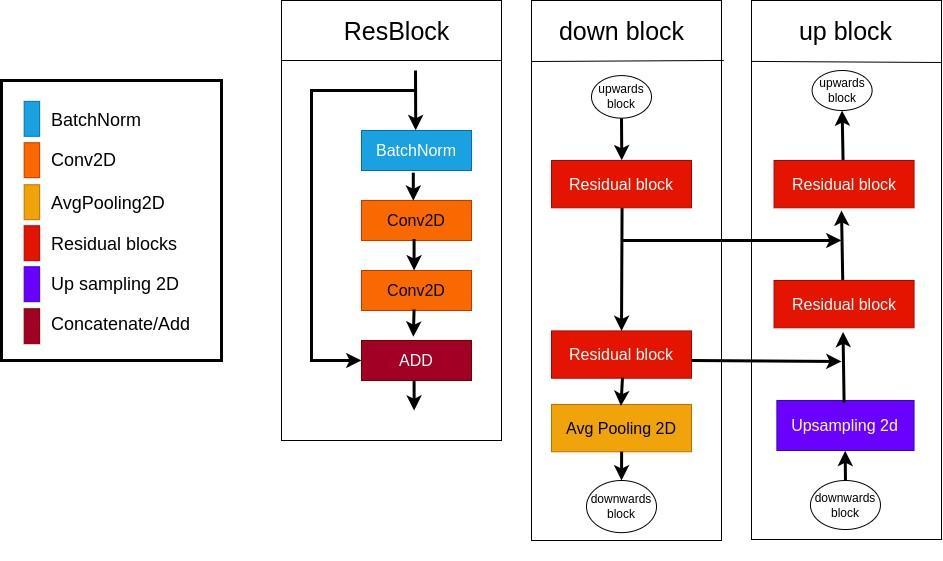}
\caption{Upsampling and Downsampling blocks consist of Residual Blocks, exploiting residual connections.
}\label{fig:components}
\end{figure}

The detailed architecture of our models is described in Fig.~\ref{fig:unet}.
Donwnsampling and upsampling blocks are composed by a short a short, configurable sequence of Residual Blocks, as described in Fig~\ref{fig:components}.

\subsection{Visual Transformer}
\label{sec:vit}
\begin{figure}[H]
\centering
\includegraphics[width=.85\textwidth]{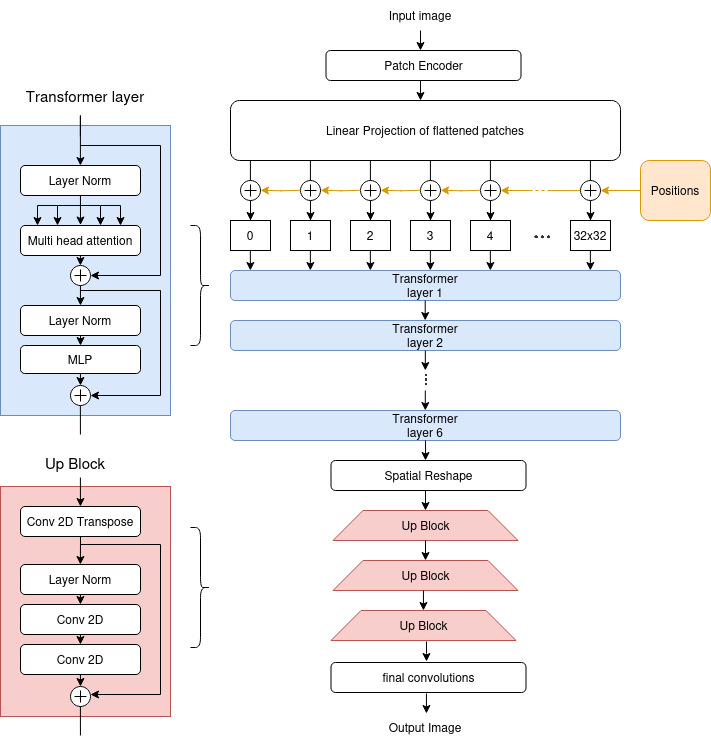}
\caption{ViT Model. The source image is divided into fixed-size patches that, after embedding, are then processed by the transformer layers. Transformer layers use multi-head self-attention to capture global dependencies across the entire image. Observe the final upsumapling blocks, peculiar to our implementation.
}\label{fig:vit}
\end{figure}

The Visual Transformer \citep[ViT;][]{dosovitskiy2020image} is a deep learning architecture designed for image recognition tasks, leveraging the transformer model \citep{vaswani2017attention}, which was originally developed for natural language processing. Unlike traditional convolutional neural networks (CNNs) that rely on convolutions to capture spatial information, ViTs use self-attention mechanisms \citep{bahdanau2015attention} 
to model the relationships between different parts of an image. Our ViT architecture is outlied in Fig.~\ref{fig:vit}.

The input image is divided into fixed-size patches, which are then flattened and projected into embeddings, similar to how words are handled in transformers for language tasks.
These patch embeddings are then processed by the transformer layers, which apply multi-head self-attention to capture global dependencies across the entire image. This approach allows ViTs to capture long-range dependencies and context more efficiently than CNNs, especially for large-scale image datasets. ViTs have demonstrated state-of-the-art performance in various vision tasks and are particularly effective when trained on large datasets. 

Relatively to the topic of cloud occlusion, a ViT model was used in \citet{goh2024maesstro}, where a random subset of SST patches is masked or removed at training time, mimicking a larger cloud occlusion. Afterward, a set of learnable mask tokens is added to the encoded patches before they are passed to the decoder, that reconstructs the original SST tile in pixel format.

In the usual ViT structure, after reassembling patches in their original spatial arrangement there is no further processing, and the output is directly produced in pixel format. However, we experimentally found convenient to add a few convolutional layers, also to recover the original spatial dimension of the input through a suitable number of upsampling operations, instead of
a mere reshaping.

\subsection{Generator for training and evaluation}
\label{sec:generator}
\begin{figure}[H]
\centering
\includegraphics[width=\textwidth]{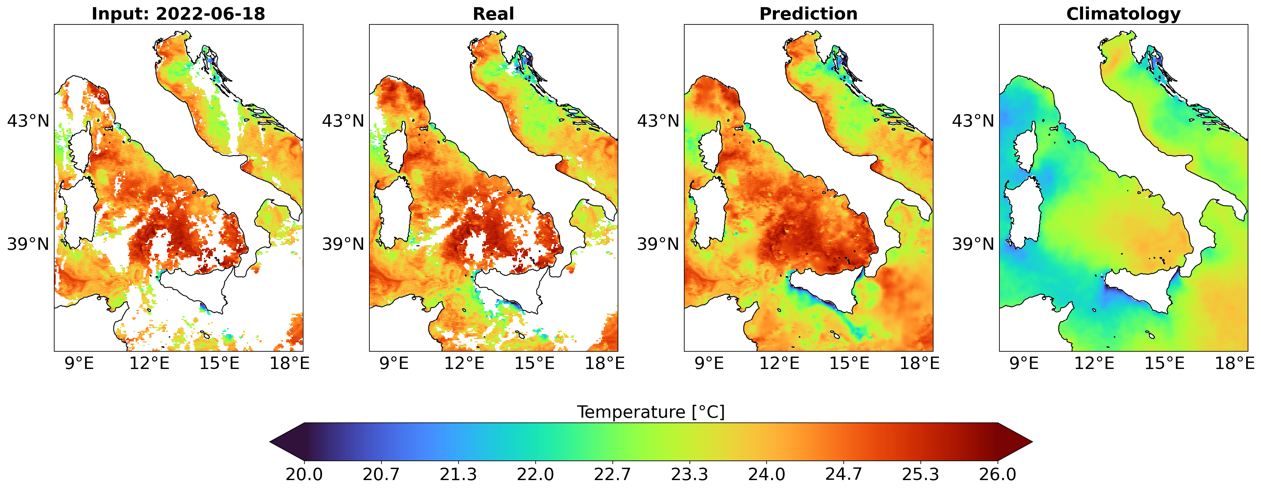}
\caption{The image on the left is the SST model input created by the generator adding artificial occlusions to the second image, which is the real or observed L3 image; the third image is the model prediction; the last on the right is the climatology.
}\label{fig:generator}
\end{figure}

The training/evaluation of the reconstruction model is not straightforward since we do not have a ground truth for comparison, i.e. we do not know the SST under the cloud on each specific night. Here we use the approach of \citet{barth2020dincae,barth2022dincae} and \citet{goh2024maesstro}, the so-called generator, which involves creating an artificial occluded area in the source image and restricting the evaluation to the region of the artificial clouds. 
The generator begins by selecting a random day from the nighttime dataset, ensuring that the chosen image has at least 40\% of the visible sea, to avoid working with insufficiently informative images. For each selected day, SST measurements relative to a given, configurable number of previous days are also retrieved. This approach enables the network to capture both spatial and temporal information, allowing the model to use historical data to reconstruct missing areas in the current image.
The clouds are selected in such a way as to guarantee a minimum percentage of visible sea (typically, 5\%), while ensuring that the artificially occluded area covers at least 10\% of the sea area. These two ranges are easily configurable. The overall procedure is meant to ensure that the image has enough occluded areas to support meaningful training and validation. Artificial masks are also applied to the previous days, maintaining temporal correlation. 
The average percentage of visible sea in nightly MODIS data relative to the Italian seas region is around 46\%. The generator produces an average visibility of around 25\%, with an average artificial cloud occlusion above 40\%. This is good for training since we expose the model to relatively challenging situations, but it is a bit unrealistic during testing. This is a delicate point since, not surprisingly, the performance of the model depends on the degree of occlusion of the input image, which becomes a crucial parameter of the evaluation. 
Starting with the real image (Fig.~\ref{fig:generator}) the generator computes the real occlusion mask. The real occluded area is changed by superimposing an artificial mask for clouds from a different day (Fig.~\ref{fig:generator}). The difference between the artificial mask and the real mask will define the region of the input where the reconstruction will be assessed. This approach risks introducing biases since the sea temperature under clouds is usually different from the temperature under a clear sky, but this bias is lower during the night. The reconstruction is given in Fig.~\ref{fig:generator}. It is interesting to observe that, despite the heavy occlusion of the input, the model can correctly reconstruct many details of the real image. Qualitative and quantitative evaluations will be given in Section~\ref{sec:results}. 

\subsection{Training}
\label{sec:training}

The models  have been developed in the Tensorflow/Keras framework and trained using the recent AdamW optimizer, which adapts the learning rate during training, combining it with weight decay. The starting learning rate was \lstinline|1e-4|.

Training was conducted with a maximum limit of 200 epochs with early stopping, using a batch size of 32. For most configurations, training stops in less than 100 epochs, due to the early stopping callback. Training exploits the generator. At the end of each epoch, a validation phase is executed, monitoring the loss on a suitable validation generator.

Various callbacks were used during training to improve model stability and performance and to perform real-time evaluations:

\begin{itemize}
    \item \textbf{Early Stopping}: Training is stopped early if the validation loss does not improve for a given number of consecutive epochs, called patience. In our case, patience was set to 10. 
    
    \item \textbf{ReduceLROnPlateau}: If the validation loss does not improve for 10 epochs, the learning rate is halved, until reaching a minimum value of \lstinline|1e-5|. This allows for smaller, more precise updates to the model in later training stages, ensuring higher performance.
    
    \item \textbf{ModelCheckpoint}: Whenever the validation loss improves, the model weights are saved to enable restoring the best model for further training.
    
    \item \textbf{TestCallbackGaussian}: A custom callback evaluates the model using RMSE at set intervals, for monitoring purposes.
    
    \item \textbf{TestCallbackGaussianDincae}: Similar to the previous callback, but the model is evaluated using DINCAE data, discussed further in Section~\ref{sec:dincae}.
\end{itemize}

\section{Model intercomparison and selection}
\label{sec:results}

This section describes the numerical experiments done with the different models and configurations to choose the optimal configuration. We primarily compare three models: two U-Nets and one ViT. The two U-Nets, referred to as U-Net32 and U-Net64, differ in the number of channels, with the latter having double the number of channels. Both models have three downsampling layers, with a topology of [32,64,128,256] for U-Net32 and [64,128,256,512] for U-Net64. The numbers 32 and 64 in the network names correspond to the initial number of channels, before downsampling. The number of parameters for the three models is provided in Table~\ref{tab:params}.

\begin{table}[h]
    \centering
    \begin{tabular}{|c|c|c|c|}
           \hline
                 & {\bf U-Net32} & {\bf U-Net64} & {\bf ViT}\\\hline
          {\bf Parameters} & 4,259,489 & 17,022,273 & 5,918,081 \\\hline
    \end{tabular}
    \caption{Number of parameters for the models. The numbers refer to the versions with 11 input channels (5 days). Shorter sequences do not notably change the total number of parameters.}
    \label{tab:params}
\end{table}

For each model, we consider two variants with different spatial dimensions: 128x128 and 256x256. The 128x128 model is a combination of four models, each trained on a different region of the input image. In this case, the evaluation metric is the average performance across the four models.
Additionally, we vary the number of "s" consecutive input days used for reconstructing the SST, from the day "t" (current day) to day "t-s". We do not consider future days in the reconstruction. 
The performance score used is the Root Mean Square Error (RMSE) calculated as the difference between the reconstructed SST under an artificial cloud occlusion and the real SST at that point. The average is measured using the test generator across a total of 50 batches, each consisting of 32 samples (for a total of 1,600 days). 
As explained in Section 3.2, the cloud occlusion generator was set to provide an average percentage of the visible sea of around 46\%, which is similar to real data. Results reported in Table~\ref{tab:rmse} show that the quality improvement given by increasing the number of days in the past, saturates after 4 days. 

\begin{table}[h]
    \centering
    \begin{tabular}{|c|c|c|c|c|c|c|}
      \hline
        {\bf days}& \multicolumn{2}{c|}{\bf U-Net32} & \multicolumn{2}{c|}{\bf U-Net64} & \multicolumn{2}{c|}{\bf ViT}\\
      {} & \scriptsize{$256\times256$} & \scriptsize{$128\times128\, (\times 4) $} & \scriptsize{$256\times256$} & \scriptsize{$128\times128\, (\times 4)$} &\scriptsize{$256\times256$} & \scriptsize{$128\times128\, (\times 4)$}\\\hline
        1 & 0.36 & 0.33 & 0.35 & 0.32 &  0.38 & 0.36\\\hline
        2 & 0.35 & 0.32 & 0.34 & 0.31 &  0.36 & 0.34 \\\hline
        3 & 0.35 & 0.31 & 0.34 & 0.30 &  0.35 & 0.33 \\\hline
        4 & 0.34 & 0.31 & 0.33 & 0.30 &  0.35 & 0.33\\\hline
        5 & 0.34 & 0.31 & 0.33 & 0.30  & 0.35 & 0.33\\\hline
        6 & 0.34 & 0.31 & 0.33 & 0.30 &  0.35 & 0.33 \\\hline
    \end{tabular}
    \caption{Performance of the different models, measured in RMSE. The different
    rows refer to the number of consecutive days passed as input to the model.
    Values refer to data with an average percentage of visible sea around $46\%$.
    Splitting the model in 4 models with lower spatial dimension consistently
    gives better results.}
    \label{tab:rmse}
\end{table}

This is consistent with our investigation of persistence (Section~\ref{sec:persistence}). On the other hand, splitting the model into smaller geographical regions of dimension 128x128 each results also increases the performance.
In Table~\ref{tab:rmse_by_quadrant} we report the details of the RMSE for the four quadrants over Italy; the values refer to our best model, namely U-Net64 with 4 days in input.

\begin{table}[h]
    \centering
    \begin{tabular}{|c|c|c|c|c|c|}
        \hline
         {\bf quadrant} &  {\bf NW} & {\bf NE} & {\bf SW} & {\bf SE} & {\bf Mean}\\\hline
         RMSE & 0.305 & 0.283 & 0.304 & 0.314 & 0.302 \\\hline
    \end{tabular}
    \caption{RMSE subdivided by quadrants. NW: North Tyrrhenian Sea, NE: Adriatic Sea, SW: South Tyrrhenian Sea, SE: Ionian Sea. The values refer to U-Net64, with 4 days in the past of input. The Adriatic is the region with the best reconstruction accuracy, while the most complex one is the Ionian Sea.
    } \label{tab:rmse_by_quadrant}
\end{table}

\begin{figure}[h]
    \centering
    \includegraphics[width=0.7\linewidth]{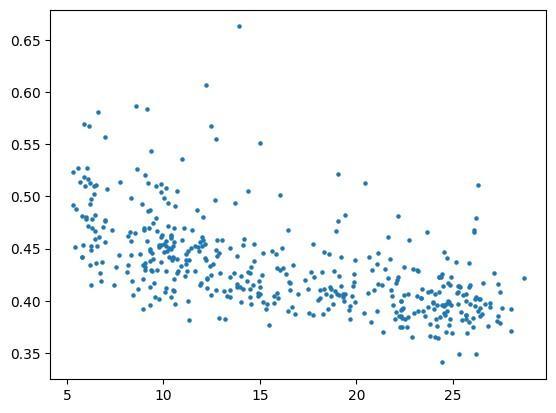}
    \caption{Degradation of the reconstruction error (vertical axis, units in $^\circ$C) with the percentage of visible sea (horizontal axis). Each point is a batch of 32 days.}
    \label{fig:RMSE_degradation}
\end{figure}

A positive feature of our configurable cloud occlusion generator is its ability to easily adjust the maximum percentage of the visible sea in the model’s input. Fig.~\ref{fig:RMSE_degradation} illustrates the relationship between visible sea percentage and RMSE, with each point representing a batch of 32 days. The plot corresponds to a U-Net32 model with 4 input days. As expected, performance degrades with higher levels of occlusion, but this degradation is nearly linear and not particularly severe. According to our investigations, there is an incremental error of approximately 0.005$^\circ$C for each additional percentage point of sea occlusion.

\subsection{Verification with DINCAE}
\label{sec:dincae}

In this section, we compare the performance of our model with that of another state-of-the-art data-driven model: the Data INterpolating Convolutional Auto-Encoder \citep[DINCAE;][]{barth2020dincae,barth2022dincae}. The comparison data are daily and at the same resolution as the data on which our model was trained, originating from nighttime SST measurements by the MODIS-TERRA satellite from 1/1/2003 to 12/31/2016. These data are divided into two datasets: one with artificially added coverage and the other with original data. Unlike our approach, the additional coverage is fixed and not configurable.
The DINCAE2 test focuses on the Northern Adriatic Sea, the analysis area differs from the one used by our model but largely overlaps. This area extends in latitude from 40$^\circ$ to 46$^\circ$ and in longitude from 12$^\circ$ to 19$^\circ$. From the model’s perspective, DINCAE \citep{barth2020dincae} is essentially a double UNet, where two networks are composed in sequence. Unlike our model, which only uses the available SST information, DINCAE incorporates additional indicators, including the date and wind speed, which are further extended in DINCAE2 \citep{barth2022dincae} to account for satellite chlorophyll.

In Table~\ref{tab:dincae_vs_our}, we compare the RMSE of DINCAE (best model, with chlorophyll) and our models with 4 days in the past as input. The U-net64 outperforms DINCAE  by approximately 20\%.

\begin{table}[h]
    \centering
    \begin{tabular}{|c|c|c|c|c|}
      \hline
        {\bf DINCAE}& \multicolumn{2}{c|}{\bf U-Net32} & \multicolumn{2}{c|}{\bf U-Net64} \\
      {\bf } & $256\times256$ & $128\times128$ & $256\times256$ & $128\times128$ \\\hline
        0.54 & 0.45 & 0.43 & 0.44 & {\bf 0.42}   \\\hline
    \end{tabular}
    \caption{Comparison of RMSE reconstruction errors (units $^\circ$C) between our 4-day input days models and DINCAE.
    }\label{tab:dincae_vs_our}
\end{table}

\section{Application of the best model to operational input data sets}
\label{sec:copernicus}

In this section we explore the applicability of our Unet-64 model with 4 days input data to the real time Copernicus Marine Core Service L3 product. The latter is a calibrated cloud occluded image at 1/16 degree resolution, merging of several available thermal imaging sensors observations. We retrained the model L3S reprocessed data between 7/4/2022 and 12/31/2023 (see Section~\ref{sec:dataset}). The reconstructions are performed using L3S NRT product as input and then we compare our reconstruction with the L4 NRT product. Our purpose is to test the best configuration against the NRT products since those are used in Copernicus Marine operational analysis and forecasting  systems. We refer to \citet{pisano2022new} for the detailed description of these datasets. 

\begin{figure}[h]
    \centering
    \includegraphics[width=0.7\linewidth]{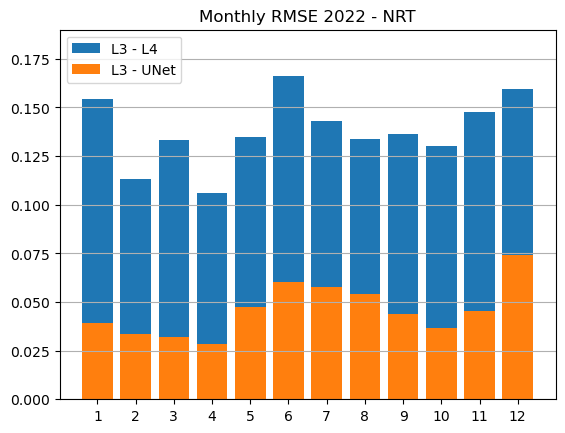}
    \caption{(Blue) Distribution over the 12 months of the year of the RMSE between L4 and L3s computed on the visible region of the sea. (Orange) Same error relative to our reconstruction.
    }\label{fig:l4_error_per_month}
\end{figure}

In Fig.~\ref{fig:l4_error_per_month}, we compare the error by the two different reconstruction methods and the L3 input over visible regions of the sea. The error is relative to the year 2022, and is shown as a function of the months. Specifically, the error of L4 is in blue, with an average RMSE of 0.14$^\circ$, while the average error of our reconstruction (in orange) is around 0.04$^\circ$. The error of the U-net64 is more uniform, and particularly low in the period from January to March.

\begin{figure}[hp]
    \centering
    \begin{tabular}{c}
         \includegraphics[width=1.\linewidth]{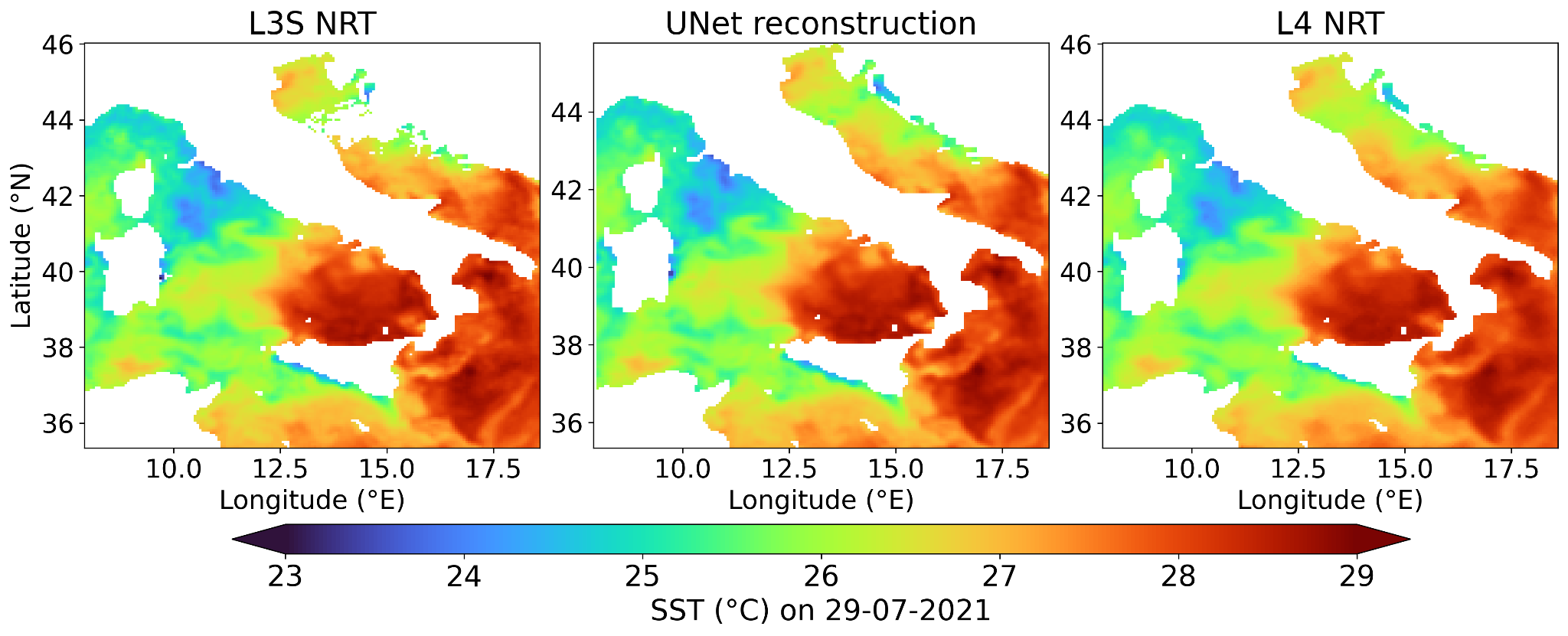} \\
         \includegraphics[width=1.\linewidth]{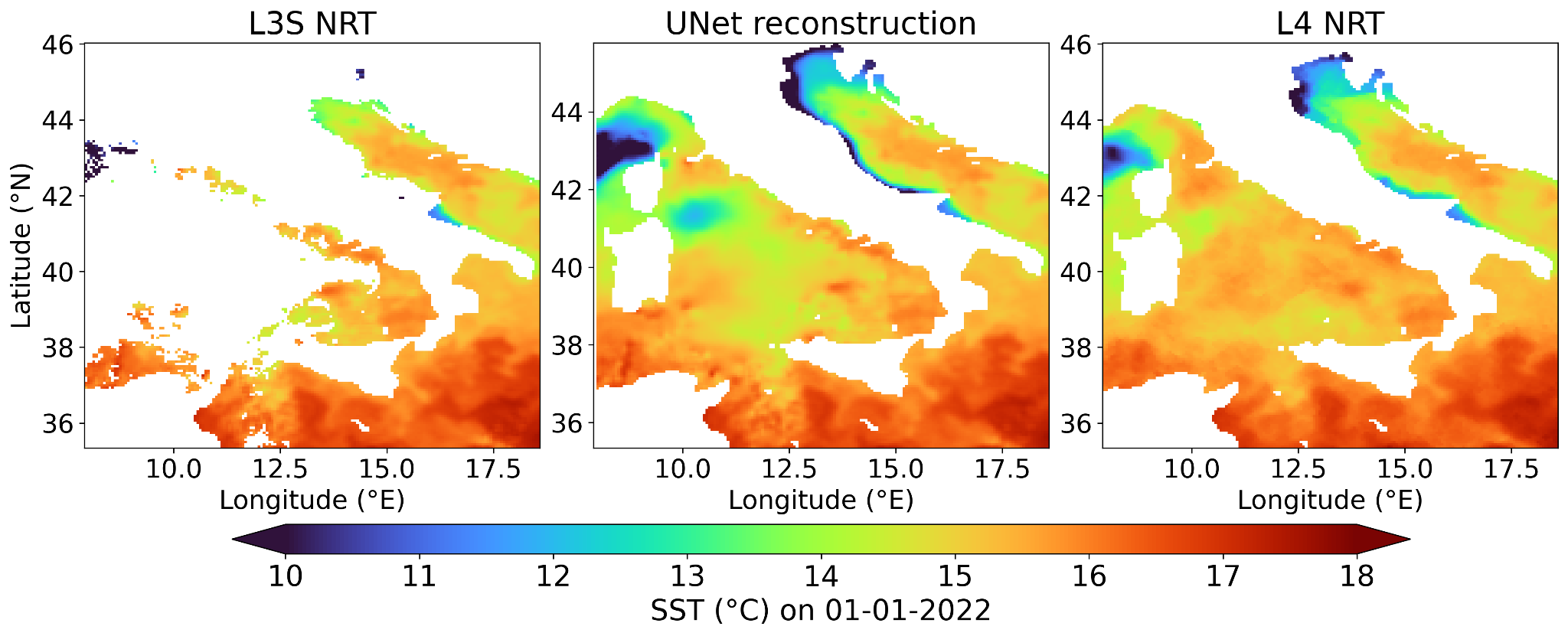} \\
    \end{tabular}
    \caption{Visual comparison between our reconstruction (U-Net32 with 4 input days) and L4 product from Copernicus Marine Service (marine.copernicus.eu) for 29 July 2021 (top) selected days and 1 January 2022 (bottom).
    }
    \label{fig:l4_visual_comparison}
\end{figure}

In Figure~\ref{fig:l4_visual_comparison} we qualitatively compare our reconstructions with those offered by the L4 product for two days in July 2021 and January 2022. Our reconstruction looks more faithful in the cloud free areas, maintaining frontal regions in a manner very similar to the original data.  We conclude that the trained network on the MODIS-AQUA data set performs very well also with different inputs data sets probably because the cloud occlusion geometry is similar in the two data sets.

\section{Discussion}
\label{sec:discussion}
Regarding the models, we tested diffusion models \citep{ho2020denoising, song2020denoising}, which we have recently applied successfully to downscaling \citep{merizzi2024wind} and precipitation nowcasting \citep{asperti2025precipitation}. Despite our efforts, we were unable to achieve competitive performance with diffusion models in this case.
We tested numerous additional variants of U-Net and ViT, tuning their architectures and incorporating more sophisticated layers. Specifically, we experimented with inception modules \citep{szegedy2015going}, Bottleneck Attention Modules \citep[BAM;][]{park2018bam}, Convolutional Block Attention Modules \citep[CBAM;][]{woo2018cbam}, and AttentionAware layers \citep{zheng2022bridging}. However, none of these mechanisms led to significant performance improvements.
From a methodological perspective, the subtraction of the unbiased climatology to compute anomalies gave us the best performance for the reconstruction of SST with artificial cloud occlusion. While the results without subtracting the unbiased seasonal cycle were acceptable, they consistently showed a performance decrease of around 10\%. In this scenario, subtracting seasonality generally led to faster and more stable training as well as improved final performance.

\section{Conclusion}
\label{sec:conclusions}

This study investigated deep neural networks to reconstruct SST data gaps caused by cloud occlusion, focusing on improving data completeness and reliability.
Our findings indicate that deep learning approaches, if properly tuned on spatial and temporal dimensions, significantly improve SST reconstruction accuracy.
Comparisons with existing methods, including L4 statistical interpolation reconstructions and other data-driven models, highlight the effectiveness of deep learning in achieving reliable SST reconstructions. Our selected best model architectures is made of a U-net64 algorithm with 3 or 4 days in the past input data and the subtraction of a long term unbiased seasonal cycle. 
In the future, we plan to extend the analysis to the full Mediterranean area and to incorporate data with microwave measurements, both for training and validation purposes. This could further enhance reconstruction accuracy, allowing models to resolve the daily cycle SST dynamics.
In summary, our results underscore the potential of deep learning to enhance SST data completeness and accuracy, offering promising applications in climate science, marine research, and oceanographic studies. 

\begin{appendices}
\section{Appendix}

The application of a Gaussian blur to a matrix with no values is not entirely straightforward. Simply replacing the no values with zeros would treat them as valid values, flattening the result. Therefore, we need to reweight the result according to the average of the filter weights corresponding to meaningful (non-zero values) locations.

The interpolation process is divided into the following stages:

\begin{enumerate}
\item Replacement of missing values. The missing values (NaN) are replaced with zeros in the data matrix.
\item Application of the Gaussian filter. A Gaussian filter is applied to the resulting matrix, creating a "blurred" (biased) version of the data.
\item Generation of a binary mask. A second matrix is created to track the originally known points, assigning a weight of 0 to points with missing values and 1 to known points.
\item Application of the Gaussian filter to the mask. The resulting weights matrix describes the actual contribution of the region under the filter.
\item Division of matrices. The blurred data matrix is divided by the weights matrix to correct the original bias.
\end{enumerate}
In the final division, the denominator may be 0, corresponding to a region in the input entirely composed of NaNs. In this case, mathematical libraries usually produce a NaN. After applying the filter, a few NaNs may still remain, which can be filled by iterating the technique or using other interpolation methods. It is worth mentioning that we interpolate both spatially and temporally to take advantage of persistence and produce a smoother climatology along the temporal axis.
The interpolation algorithm used to create a climatology from satellite SST nighttime values should account for the fact that certain grid points consistently lack values due to cloud coverage and coastal, shallow water low temperatures. Therefore, an extrapolation procedure needs to be developed. In this study, the following code, written in NumPy style, was employed.

\begin{algorithm}
\caption{Gaussian Blur Interpolation with NaNs}\label{alg:gaussian_blur}
\begin{algorithmic}[1]
\State \textbf{Input:} Data matrix $D$ with NaNs, Gaussian filter $G$
\State \textbf{Output:} Interpolated matrix $D_{interpolated}$ 

\State $D_{copy} = D.copy()$ \Comment make a copy of $D$
\State $D_{zeroed} = D_{copy}[np.isnan(D)]=0$ \Comment replace NaN with zero
\State $D_{blurred}= gaussian\_filter(D_{zeroed},G)$ \Comment apply $G$ to $D_{zeroed}$
\State $M=np.where(np.isnan(D),0,1)$ \Comment create a bynary mask
\State $W= gaussian\_filter(M,G)$ \Comment apply $G$ to the mask $M$
\State $D_{interpolated} = D_{blurred}/W$ \Comment correct the bias 
\end{algorithmic}
\end{algorithm}
\end{appendices}

\section*{Code and Data} The code developed in this work is available in the GitHub repository at the following url: \url{https://github.com/asperti/SST_reconstruction}. Main datasets may be accessed trough the notebooks in the repository or downloaded from the sites specified in the work.

\section*{Acknowledgements}
This research was partially funded and supported by the following Projects:
\begin{itemize}
\item Future AI Research (FAIR) project of the National Recovery and Resilience Plan (NRRP), Mission 4 Component 2 Investment 1.3 funded from the European Union - NextGenerationEU.
\item ISCRA Project ``AI for weather analysis and forecast'' (AIWAF2)
\item CMEMS Med-MFC (Copernicus Marine Service - Mediterranean Sea Marine Forecasting Center), Mercator Ocean International.
\end{itemize}


\end{document}